\documentclass[preprints,article,accept,pdftex,moreauthors]{mdpi} 
\firstpage{1} 
\pubvolume{1}
\issuenum{1}
\articlenumber{0}
\pubyear{2026}
\copyrightyear{2026}
\datereceived{ } 
\daterevised{ }
\dateaccepted{ } 
\datepublished{ } 

\Title{Validation of Smartphone-Based Photogrammetric 3D Body Scanning for Automated Anthropometric Measurements Compared with a Commercial Depth-Sensor-Based Body Scanner}

\Author{Ruting Cheng $^{1}$*\orcidA{}, Boyuan Feng $^{1}$\orcidB{}, Chuhui Qiu $^{1}$\orcidC{}, Joaquin A. Calderon $^{2}$\orcidD{}, Qing Pan $^{3}$\orcidE{}, Yufan Liu $^{4}$\orcidF{} and James K. Hahn $^{1}$\orcidG{}}

\AuthorNames{Ruting Cheng, Boyuan Feng, Chuhui Qiu, Joaquin A. Calderon, Qing Pan, Yufan Liu and James K. Hahn}

\address{%
$^{1}$ \quad Department of Computer Science, The George Washington University, Washington DC, 20052, USA; fby@gwu.edu (B.F.); chqiu@gwu.edu (C.Q.); hahn@gwu.edu (J.H.)\\
$^{2}$ \quad Department of Obstetrics and Gynecology, The George Washington University, Washington DC, 20052, USA; joacocal91@gmail.com;\\
$^{3}$ \quad Department of Biostatistics and Bioinformatics, The George Washington University, Washington DC, 20052, USA; qpan@gwu.edu;\\
$^{4}$ \quad School of Computing and Data Science, The University of Hong Kong, Hong Kong SAR; yufan.liu@connect.hku.hk}

\corres{Correspondence: rcheng77@gwu.edu;}

\abstract{3D body scanning has become an important tool in healthcare applications because of its rapid and non-invasive nature. While smartphone-based photogrammetric reconstruction provide a low-cost and accessible alternative to commercial 3D body scanners, their performance for whole-body scanning remains insufficiently validated. Thus, we designed this study to comprehensively validate the photogrammetric 3D scanning application by evaluating automatically extracted whole-body measurements and longitudinal body-shape monitoring. We evaluated a representative application, PolyCam, against the commercial depth-sensor-based Fit3D ProScanner using 144 pregnant participants scanned longitudinally throughout pregnancy. We designed an automatic circumference extraction pipeline to get measurements at four anatomical landmarks from paired 3D scans. A linear mixed-effects model was used to evaluate scanner effects and longitudinal body-shape changes. Measurement consistency was assessed using repeated PolyCam scans and tape measurements on a rigid mannequin. PolyCam demonstrated strong agreement with Fit3D, with average biases below 16 mm, intraclass correlation coefficients above 0.8, and Pearson correlation coefficients above 0.9 across all landmarks. Both systems captured comparable longitudinal body-shape changes. Mannequin experiments showed mean biases below 3.5 mm and no significant differences from tape measurements. These findings support smartphone photogrammetry as a potential accessible alternative to commercial body scanners and applicable for longitudinal 3D body-shape assessment.} 

\keyword{3D body scanning, photogrammetric reconstruction, smartphone-based scanning, digital health} 

\begin{document}

\section{Introduction}
With the development of 3D scanning technology, 3D body surface scanning is becoming prevalent not only in the fashion and gaming field, but also in the medical and clinical domains \cite{kim2024accuracy,pei2019female, jung2022impact,minetto2026feasibility, feng2025enhanced, balasubramanian2025emerging, brownridge2014body}. Compared with traditional anthropometric measurements, 3D body shape provides substantially richer geometric information while demonstrating considerable value in body composition estimation, nutritional assessment, and the identification of obesity-related health conditions\cite{ng2016clinical, tian20253d, feng2026voxel, zheng2024d3bt, zheng2024predicting, poltronieri2026using}. More recently, 3D body shape analysis has shown promise in obstetrics for fetal growth estimation and cesarean delivery prediction\cite{cheng2026maternal, dathan2023novel, gleason2018safe, cheng2025mvbody}, in orthopedics for scoliosis assessment\cite{oquendo2025mobile, kuo2026comparison, roy2019noninvasive}, and in sports science for training optimization\cite{simenko2017body, schueler2018using, vsimenko2024rapid}. In addition, the ability of 3D scanning to preserve both surface geometry and color information has attracted growing interest in medico-legal applications \cite{callegari2024precision, kottner2023using}.

The widespread adoption of 3D body scanning in medical and home-based settings depends largely on its affordability, accessibility, safety, and ease of operation. However, most existing studies rely on commercial depth-sensor-based body scanners, such as Fit3D ProScanner (Fit3D, San Francisco, CA, USA) and Styku (Styku LLC, Los Angeles, CA, USA). Despite their high accuracy, these commercial body scanners remain relatively expensive and are typically limited to specialized clinical or research facilities, restricting their broader adoption for community-based or home-based monitoring \cite{ng2016clinical, feng2026voxel, cheng2026maternal, ashby2023translating, dalamaggas2022assessing}. Alternative portable solutions have been explored, which can be broadly categorized into two types. The first category employs specific type of devices or general devices equipped with dedicated depth sensors. For example, Callegari et al. evaluated the accuracy of PolyCam (Polycam Inc., San Francisco, CA, USA) running on a LiDAR-equipped smartphone for medico-legal body documentation\cite{callegari2024precision}. Similarly, Oquendo et al. integrated a Structure Sensor Mark II (Occipital Inc., Boulder, CO, USA) with an iPad Pro (Apple Inc., Cupertino, CA, USA) and demonstrated its feasibility for adolescent idiopathic scoliosis screening by capturing 3D trunk morphology\cite{oquendo2025mobile}. Without requirement of special hardware, the second category of solutions for portable body-scanning is template-based reconstruction. Poltronieri et al. evaluated a representative application, named MeThreeSixty (Size Stream LLC, Cary, NC, USA), which estimates 3D body shape by fitting a small number of 2D images to a predefined human body template\cite{poltronieri2026using}. Idrees et al. evaluated another two applications, Mobile Fit (Size Stream, Cary, NC, USA) and Prism Labs (Prism Labs, Los Angeles, CA, USA), which use two photos and a series of photos captured during the user's full 360° rotation separately, for parameterized body model fitting \cite{idrees2024mobile}. Although this approach is computationally efficient and highly accessible, the reconstructed body shape is constrained by the underlying template model and may fail to capture fine anatomical details and individual body-shape variations. These limitations may become more pronounced for populations with atypical body shapes, such as pregnant individuals, people with severe obesity, and amputees with limb loss \cite{aiersilan2025investigating}. Consequently, there is a need for reconstruction approaches that can accurately preserve geometric fidelity without relying on dedicated hardware or predefined body models.

Unlike these two kinds of methods that use depth sensors for 3D body shape capturing or parameterize 3D body shape with templates, photogrammetric 3D reconstruction offers an alternative approach for affordable 3D body scanning with high accuracy. By identifying corresponding features across multiple images and reconstructing their spatial locations through triangulation, photogrammetry can generate detailed 3D surface models without relying on specialized sensors or predefined body templates. Recent studies have evaluated popular photogrammetric smartphone applications in medical settings. Rudari et al., evaluated 3 professional 3D scanner and 2 3D scanning applications, Scandy Pro (Scandy LLC, New Orleans, LA, USA) and PolyCam, for human hand scanning\cite{rudari2024accuracy}. Walters et al. compared 3 smartphone applications, PolyCam, Luma (Luma AI Inc, Palo Alto, CA, USA), and Meshroom (AliceVision, Paris, France), to a professional 3D scanner Artec Eva (Artec 3D, Luxembourg City, Luxembourg) on residual limb scanning for health monitoring and socket design\cite{walters2024smartphone}. Both studies reported favorable performance of photogrammetry-based reconstruction, with PolyCam demonstrating superior performance among the evaluated smartphone applications. Despite these encouraging findings, existing validation studies have almost exclusively examined localized anatomical regions, such as the hand or residual limb, on relatively small datasets. Whether smartphone-based photogrammetric reconstruction can achieve comparable accuracy for complete human body acquisition remains largely unknown. Whole-body scanning introduces additional challenges, including subject motion, changing viewpoints, and reconstruction over substantially larger anatomical surfaces, all of which may influence measurement accuracy.

Therefore, this study aimed to evaluate the accuracy and consistency of a photogrammetric smartphone approach for whole-body 3D scanning under both realistic scanning conditions using human participants and controlled conditions using a rigid mannequin. Regarding the human dataset, we collected longitudinal 3D body shapes of pregnant participants, which involves both realistic body characteristics and progressive anatomical changes. Circumference measurements automatically extracted from photogrammetric reconstructions were compared with those obtained from a commercial depth-sensor-based scanner. The ability of both scanning approaches to capture longitudinal body-shape changes was evaluated using scans acquired from pregnant participants at different stages of pregnancy. To assess the consistency of the photogrammetric approach, repeated scans of a rigid mannequin were performed and compared with reference tape measurements, thereby eliminating variability introduced by physiological motion. We hypothesized that (1) the photogrammetric approach would demonstrate high agreement with measurements obtained from the commercial scanner, with small and generally acceptable bias, and (2) the photogrammetric approach would exhibit good repeatability and consistency on the mannequin, with lower variability than that observed in human participants. Representative products from each technology category were selected for evaluation, namely PolyCam, as a photogrammetric reconstruction application and Fit3D ProScanner as a commercial depth-sensor-based body scanner. The selection was based on market rating, our preliminary testing and evidence reported in previous studies \cite{rudari2024accuracy, walters2024smartphone, ng2016clinical, sobhiyeh2021digital}.

To our knowledge, this is the first study to systematically evaluate a smartphone-based photogrammetric application for whole-body 3D scanning. The findings are expected to inform the development and adoption of smartphone-based photogrammetric whole-body 3D reconstruction technologies for clinical research, remote health monitoring, and future computational body-shape analysis.

\section{Materials and Methods}
\subsection{3D Scanning Device}
PolyCam (Polycam Inc., San Francisco, CA, USA) was chosen as the representative photogrammetry-based application based on its good performance in our preliminary testing and reports from previous validation studies \cite{rudari2024accuracy, walters2024smartphone}. And two smartphone models, iPhone 13 and iPhone 14 (Apple Inc., Cupertino, CA, USA), are used for scanning. In its image mode, PolyCam reconstructs 3D surface geometry from multiple overlapping RGB images using a photogrammetric reconstruction pipeline based on feature matching and multi-view geometry. Depending on scene complexity, reconstruction typically requires approximately 20–2000 RGB images. PolyCam supports both fixed position scanning, in which the target is rotating on a turntable, and dynamic scanning, in which the target keeps standing still and the operator takes photos around the target from different viewpoints. In our experiments, "object" option is selected before scanning. "Photogrammetry", "dense" resolution and "isolate object from environment" options are selected in the interface before processing. In addition, image sets can be uploaded through the PolyCam web platform, enabling centralized processing for applications such as community-based healthcare and facilitating large-scale data collection.

Fit3D ProScanner (Fit3D Inc., San Francisco, CA, USA) was selected as the representative commercial depth-sensor-based scanner due to the justified high precision and repeatability for anthropometric measurements obtained from this device \cite{ng2016clinical,sobhiyeh2021digital}, and its wide adoption in medical researches \cite{minetto2022digital, cheng2026maternal, wong2019children}. The Fit3D ProScanner employs three vertically aligned depth cameras, two handles adjustable along vertical direction for maintaining a standardized posture, and a motorized turntable to acquire complete body surface geometry during a 30-second scan.

\subsection{3D Optical Scan of Human Participants}
\subsubsection{Recruitment and Data Collection}
In this study, we used a dataset collected as part of an obstetrics research project investigating the use of 3D body shape for prenatal care. The dataset includes 3D body scans acquired at two stages of pregnancy using both the Fit3D ProScanner and PolyCam in image mode. In this dataset, A total of 144 pregnant participants were enrolled. Exclusion criteria included: (1) age under 18; (2) multiple gestations; (3) presence of an enlarged fibroid uterus; (4) BMI greater than 60; (5) presence of any unstable medical or emotional condition or chronic disease that could interfere with study participation; (6) history of body altering procedures such as liposuction or plastic surgery. 

The first visit was scheduled between 18 and 24 weeks of gestation, and the second visit was scheduled between 31 and 38 weeks of gestation. Each participant underwent two to three scanning sessions per visit using both PolyCam and Fit3D simultaneously. Before starting the scan, participants were instructed to wear form-fitting clothing and tightly secure their hair. During scanning, participants were instructed to hold the scanner handles and maintain an “A-pose” throughout the scanning process. The Fit3D turntable rotated automatically while the vertically aligned 3 depth cameras to fully capture the body shape from 360 degrees. While the Fit3D scanner was acquiring the body shape through turntable rotation, the operator simultaneously captured images using PolyCam in Photo Mode with a smartphone mounted on a tripod positioned 2.5 m from the participant. A sequence of images was captured during the rotation of the participant to reconstruct the 3D body surface. Each scan lasted approximately 30 seconds, during which about 35 images were captured. Three participants were excluded because all scans acquired by either Fit3D or PolyCam had significant distortion. Finally, with the highest-quality scan from each visit selected for analysis, 228 scans from the 141 participants were collected in total, including 116 scans acquired during the first visit and 112 scans acquired during the second visit. 

\subsubsection{Human Scan Data Pre-processing}
\begin{figure}[H]
\isPreprints{\centering}{}
\includegraphics[width=\textwidth]{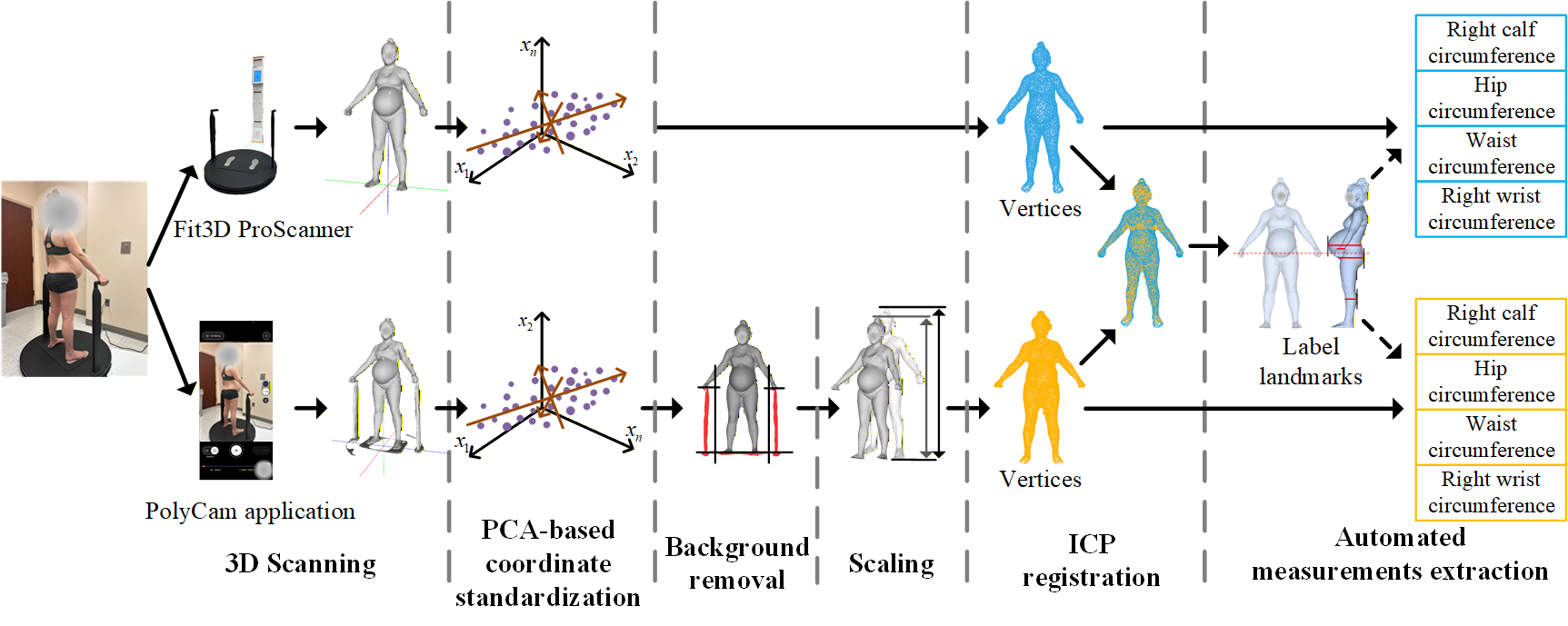}
\caption{Pre-process of the 3D human body shapes scanned by Fit3D ProScanner and PolyCam application.\label{process}}
\end{figure}

Unlike 3D scans obtained from the commercial 3D body scanner, raw PolyCam reconstructions may contain surface noise, scaling inconsistencies, and non-human structures such as scanner handles and the turntable. Thus, before extracting measurements from the scans, the 3D body scans obtained using Fit3D ProScanner and PolyCam application were pre-processed to get clean and aligned pairs. The pre-process operations are illustrated in Figure~\ref{process}.\footnote{The code will be released after the paper is accepted.}

\textbf{Coordinate standardization }
Because Fit3D and PolyCam scans are generated in different coordinate systems, mesh alignment was required prior to analysis. To standardize the coordinates of meshes from different sources, we first centered the mesh, then transformed the mesh using standardization matrix calculated from Principal Component Analysis (PCA) results with the following steps. 

Given the coordinates of the extracted point cloud of 3D body mesh, the dominant anatomical axes of each mesh, represented as principal components capturing the largest variation in the data, can be estimated using PCA. The top 3 principal components correspond to the vertical, mediolateral, and anterior directions of the 3D body scan, which can be represented as a $3\times3$ matrix $C$, with extracted top 3 principal components as row vectors. Prior to analysis, each scan is transformed to a canonical orientation such that the subject's superior–inferior, mediolateral, and anteroposterior anatomical axes are aligned with the z-, x-, and y-axes, respectively, with the anterior direction corresponding to the positive y-axis. This target orientation can be represented as matrix $T$, where
\begin{linenomath}
\begin{equation}
T = \begin{bmatrix}
    0  & 0 & 1 \\
    1  & 0 & 0 \\
    0  & 1 & 0
\end{bmatrix}.
\end{equation}
\end{linenomath}
Suppose the transformation matrix is denoted as $R$, the relationship of matrices can be represented as:
\begin{linenomath}
\begin{equation}
RC^T = T^T.
\end{equation}
\end{linenomath}
Since the principal components are orthonormal, the transformation matrix $R$ can be calculated by:
\begin{linenomath}
\begin{equation}
R = T^TC.
\end{equation}
\end{linenomath}
By applying the calculated $R$ to the whole meshes, the orientations of all meshes are standardized into a common coordinate system. For the few upside down or facing-back outputs, manual flipping is needed. Both Fit3D scans and PolyCam scans need to go through this coordinate standardization process.

\textbf{Non-human parts removal and scaling of PolyCam scans }
Since PolyCam app is not designed specifically for human scan, the turntable and scanner handles may be reconstructed together with the body surface. These non-human parts need to be removed before target measurements extraction. The mesh standardization allows critical points to be easily located for segmentation and scaling. As shown in Figure~\ref{process}, to remove the turntable, the plane between feet is detected and all mesh components below this plane were removed. For the handle removal, we use critical points that have extreme $x$ values of the body scan to locate the handles' positions. The top points of handles can be located by indexing the minimum and maximum $x$ values of the scan. Given the $x$ values and averaged $z$ value of the two handles tops, the handles can be segmented and removed. Additionally, the PCA-based coordinate standardization is repeated again after the turntable is removed to further adjust the 3D body scan for more accurate handles location.

After non-human parts are removed, the PolyCam mesh was scaled to match the Fit3D counterpart. In case there are vertices of feet removed with turntable, we keep using inter-handle distance for scaling. As the scale ratio is not constant across scan pairs, the PolyCam mesh for each pair was scaled by the ratio of the corresponding Fit3D scan width to the PolyCam scan width to align its scale with the matched Fit3D scan.

\textbf{Registration of PolyCam captured mesh to Fit3D captured mesh }
The preceding steps provided an initial alignment between the PolyCam and Fit3D meshes. Fine registration was subsequently performed on the vertices of meshes from two sources using the Iterative Closest Point (ICP) algorithm. Given a source point set $P={p_i}$ and a target point set $Q={q_i}$, the ICP algorithm iterates over the two steps: (1) establishing correspondences by finding the closest point $q_i\in Q$ for each $p_i$, where $P$ is transformed with current $S$; (2) updating the optimal transformation $S$ by minimizing the following objective
\begin{linenomath}
\begin{equation}
E(S) = \sum_{i}{\|Sp_i-q_i\|^2}.
\end{equation}
\end{linenomath}
Before applying ICP to the meshes, the vertices of meshes are downsampled to 10000 points using Farthest Point Sampling (FPS) algorithm. The downsampled vertices of PolyCam meshes are set as source point set and the downsampled vertices of Fit3D meshes are set as target point set. After applying ICP to the corresponding point sets, the optimal transformation $S$ was calculated. The resulting transformation matrix $S$ was subsequently applied to the full-resolution PolyCam mesh to obtain the final registered reconstruction.  

\textbf{Automated measurements extraction } \label{sec landmark}
With the steps described above, the meshes have non-human parts removed and are on a standard coordinate, while the meshes still have noise. We adopt built-in "Merge Close Vertices" and "Laplacian Smooth" functions in MeshLab (Visual Computing Laboratory, Pisa, Italy) to smooth the mesh surfaces. Four geometry-driven landmarks were then defined for automated circumference extraction. Rather than relying on conventional anatomical landmarks that require manual identification and are prone to human error, measurement locations were defined using automatically detectable geometric extrema on the body surface. Circumference measurements were obtained from horizontal cross-sections passing through the corresponding landmark heights. This strategy improves measurement reproducibility and facilitates large-scale automated body-shape analysis.

The four automatically detected landmarks were defined as follows: (1) on the right calf at the height of the maximum lateral protrusion of the right calf in the sagittal view; (2) on the hip at the height of the maximum posterior protrusion of the buttocks in the sagittal view; (3) on the waist at the height of the maximum anterior protrusion of the abdomen in the sagittal view; (4) on the right wrist located 70 mm above the widest part of the hand–handle junction on the right side. All positions were defined and acquired based on Fit3D scans. Since the experiment is only for accuracy evaluation, all values are measured parallel to the ground for simplification.

\subsection{Repeated Polycam Scans of Rigid Mannequin}
In addition to the human study, which evaluated PolyCam under realistic whole-body scanning conditions through simultaneous PolyCam and Fit3D acquisitions, we also assessed the repeatability and consistency of PolyCam using a rigid mannequin. This experiment focused on the performance of the scanning approach itself by minimizing variability arising from subject movement and physiological motion. Before scanning, we marked the four landmarks of right wrist, waist, hip, and right calf with markers, which are slightly different from the definition described in Section~\ref{sec landmark} due to the mannequin's shape and pose. And we fixed the mannequin in an indoor environment with moderate artificial lighting. The mannequin is scanned by an operator holding a smartphone and capturing photographs from multiple viewpoints while moving around the mannequin. The scanning is repeated 15 times. To mimic realistic user behavior and variations in scanning conditions, we varied camera-to-subject distance and number of acquired images in different scanning rounds. Images were captured at distances ranging from 0.7 m to 2.3 m, and the numbers of photos ranging from 20 to 62 with scan duration varying from 50s to 125s. For PolyCam scans on rigid mannequin, only scaling is needed for pre-processing. Circumferences are digitally measured on MeshLab. Tape measurements are also repeated 15 times on the four labeled landmarks for comparison.

\subsection{Statistical Analysis}
\subsubsection{Accuracy Analysis on Human Data}
To evaluate the effects of scanner type while accounting for gestational stages, a linear mixed-effects model (LMM) was employed. Scanner type and visit were included as fixed effects, whereas participant ID was included as a random effect to account for within-subject correlation arising from measurements of each participants in two visits. The model was specified as:
\begin{linenomath}
\begin{equation}\label{LMM}
Y = \beta_0 + \beta_1 Scanner Type + \beta_2 Visit + u + e,
\end{equation}
\end{linenomath}
where $Y$ is the measurements, $\beta_0$ is the global intercept, $\beta_1$ and $\beta_2$ are fixed-effect slopes for scanner type and order of visit, $u$ is the random intercept for patient ID, and $e$ is the residual error.

Circumferences on the 4 landmarks were gathered and analyzed separately. The bias between Fit3D measurements and PolyCam measurements were summarized. Intraclass Correlation Coefficient (ICC) was used to evaluate the reliability and agreement between scanning methods. Root Mean Square Error (RMSE) and Mean Absolute Error (MAE) were calculated on the measurement pairs to quantify how different the measurements from two scanners are. Pearson correlation coefficients were calculated to present the linear association between two scanning methods, and the Wilcoxon signed-rank test was employed to assess whether significant differences existed in the medians of measurements from two scanning methods.  

To visualize the scanning results, the Bland-Altman method was used, showing the agreement and systematic bias between the measurements obtained from Fit3D scanner and PolyCam. Scatter plots were provided to show how PolyCam measurements and Fit3D measurements aligned on different landmarks. Heatmaps of vertex-level deviations were generated on 3D body scan for qualitative analysis, which was realized using Open3D library with Python, showing the distance between each vertex of PolyCam scan and its closest vertex from the corresponding Fit3D scan when overlapping. 

\subsubsection{Consistency Analysis on Mannequin Data}
As for consistency evaluation, we applied Mann-Whitney U test to compare the repeated tape measurements and PolyCam measurements and determine if there are significant differences between two independent groups. Besides basic statistical measures such as mean and standard deviation, coefficient of variation (CV) was also calculated to quantify relative measurement variability across repeated scans. For results visualization, Box plot was used to illustrate how the measurements obtained by PolyCam and tape distribute.

\section{Results}
\subsection{Accuracy Evaluation}
Summary statistics of circumference measurements extracted from 228 paired PolyCam and Fit3D scans are presented in Table~\ref{stat}. Measurements were additionally stratified by visit to evaluate changes across gestation. The measurements obtained during the second visit were observed larger than those obtained during the first visit. Four waist measurement pairs and three hip measurement pairs were excluded because collapsed or merged body parts in either the Fit3D or PolyCam scan prevented reliable circumference extraction.

\begin{table}[H] 
\caption{Summary statistics of circumference measurements obtained from Fit3D and PolyCam at the four landmarks, with measurements from the 2 different visits discussed together and separately. \label{stat}}
\begin{tabularx}{\textwidth}{lllllllll}
\toprule
          & \multicolumn{4}{c}{Fit3D (mm)}           & \multicolumn{4}{c}{PolyCam (mm)}            \\
\cmidrule(r){2-5} \cmidrule(l){6-9}
          & mean    & std    & min    & max     & mean    & std    & min    & max     \\
\midrule
calf (N=228) & 397.81  & 41.61  & 243.04 & 502.34  & 382.80  & 44.57  & 214.05 & 490.08  \\
calf-1 (N=116) & 393.32  & 43.60  & 243.04 & 502.34  & 377.64  & 47.26  & 214.05 & 490.08  \\
calf-2 (N=112) & 402.47  & 38.89  & 278.25 & 494.88  & 388.13  & 40.93  & 253.99 & 486.14  \\
hip (N=225) & 1108.76 & 107.66 & 877.66 & 1494.88 & 1095.84 & 106.35 & 857.57 & 1433.95 \\
hip-1 (N=116) & 1089.84 & 98.29  & 936.47 & 1377.35 & 1076.31 & 99.25  & 911.07 & 1318.61 \\
hip-2 (N=109) & 1128.89 & 113.40 & 877.66 & 1494.88 & 1116.62 & 109.66 & 857.57 & 1433.95 \\
waist (N=224) & 1064.23 & 117.68 & 835.19 & 1363.60 & 1056.42 & 118.60 & 827.16 & 1365.36 \\
waist-1 (N=112) & 1005.85 & 99.78  & 835.19 & 1311.44 & 996.88  & 101.90 & 827.16 & 1326.09 \\
waist-2 (N=112) & 1122.62 & 104.53 & 861.74 & 1363.60 & 1115.96 & 103.24 & 854.46 & 1365.36 \\
wrist (N=228) & 193.50  & 23.80  & 150.50 & 273.73  & 183.28  & 22.22  & 142.78 & 263.04  \\
wrist-1 (N=116) & 191.67  & 25.09  & 162.43 & 273.73  & 180.69  & 22.23  & 148.42 & 248.28  \\
wrist-2 (N=112) & 195.39  & 22.23  & 150.50 & 260.59  & 185.96  & 21.88  & 142.78 & 263.04  \\
\bottomrule
\end{tabularx}
\noindent{\footnotesize{Note: The "-1" or "-2" after location names means this group of measurements are from scans obtained in participants' 1st visit between 18-24 gestational weeks or the 2nd visit between 31-38 gestational weeks. $N$ in the bracket denotes the number of samples. }}
\end{table}

Table~\ref{stat} shows that circumference measurements increased between visits at all four anatomical landmarks, with the largest increase observed for waist circumference. PolyCam consistently yielded slightly smaller measurements than Fit3D while exhibiting comparable variability across landmarks.

With the extracted measurements, we applied LMM to evaluate how the scanning methods related to the measurements. Outputs of LMM on different landmarks are listed in Table~\ref{LMMoutput}.

\begin{table}[H] 
\caption{LMM output of fixed effects.\label{LMMoutput}}
\isPreprints{\centering}{}
\begin{tabular}{llllll}
\toprule
Parameter   & Estimate & Std.Error & df      & t-value & Pr(\textgreater{}|t|) \\
\midrule
\textbf{Right calf} & & & & &\\
(Intercept) & 392.032  & 3.716     & 164.516 & 105.493 & \textless 2e-16       \\
scannerpoly & -15.018  & 1.436     & 312.228 & -10.459 & \textless 2e-16       \\
Visit2      & 10.731   & 1.627     & 324.352 & 6.594   & 1.74E-10 \\
\midrule
\textbf{Hip} & & & & &\\
(Intercept) & 1088.635 & 8.833     & 146.688 & 123.246 & \textless 2e-16       \\
scannerpoly & -12.919  & 1.808     & 307.068 & -7.146  & 6.53E-12              \\
Visit2      & 40.444   & 2.086     & 310.561 & 19.387  & \textless 2e-16 \\
\midrule
\textbf{Waist} & & & & &\\
(Intercept) & 1006.425 & 8.785     & 150.297 & 114.556 & \textless 2e-16       \\
scannerpoly & -7.815   & 2.288     & 305.895 & -3.416  & 0.000721              \\
Visit2      & 115.62   & 2.629     & 311.527 & 43.974  & \textless 2e-16 \\
\midrule
\textbf{Right wrist} & & & & &\\
(Intercept) & 190.954  & 1.979     & 203.333 & 96.469  & \textless 2e-16       \\
scannerpoly & -10.215  & 1.17      & 310.897 & -8.729  & \textless 2e-16       \\
Visit2      & 5.468    & 1.307     & 340.425 & 4.185   & 3.63E-05 \\
\bottomrule
\end{tabular}
\end{table}

\begin{figure}[H]
\isPreprints{\centering}{} % Only used for preprints
\includegraphics[width=0.75\textwidth]{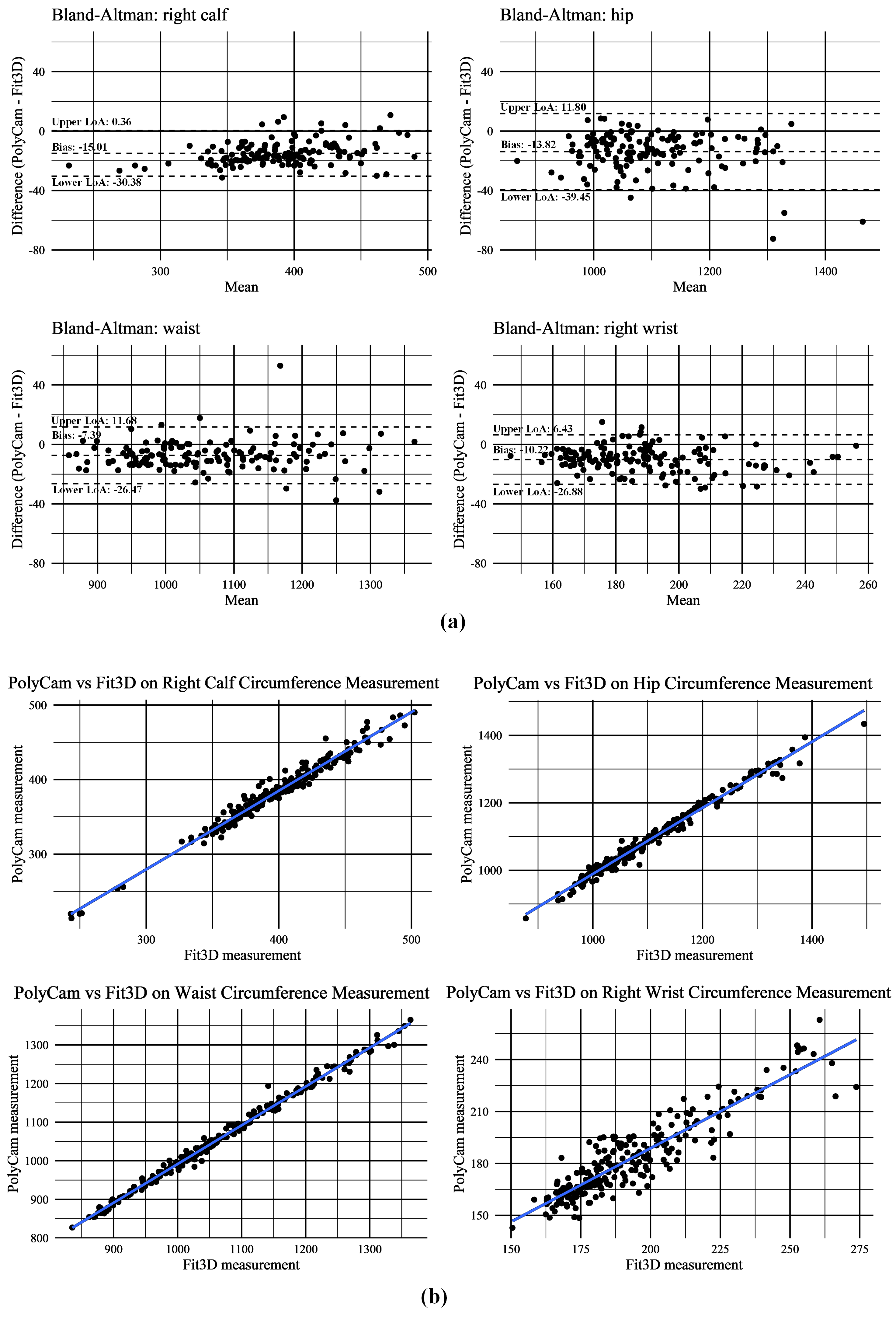}
\caption{Comparison of circumference measurements obtained from Fit3D and PolyCam at the four landmarks. (a) Bland-Altman plots, where dashed lines represent limits of agreement and solid line represents mean bias; (b) scatter plots.\label{B-A&Scatter}}
\end{figure}

In Table~\ref{LMMoutput}, "Estimate" is the estimation of $\beta$ described in Equation~\ref{LMM}; standard error, degrees of freedom (df) and t-value are used to estimate $p$ value. The LMM results provided in Table~\ref{LMMoutput} were consistent with the descriptive statistics presented in Table~\ref{stat}: (1) both scanner type and visit order were significantly associated with measurements at all four landmarks; (2) PolyCam measurements were consistently lower than Fit3D measurements; (3) Measurements obtained during the second visit were significantly larger than those obtained during the first visit on all 4 landmarks particularly on the waist. Focusing on the influences of scanning methods, further analysis was conducted on the bias between PolyCam and Fit3D scan pairs (PolyCam - Fit3D), of which the results are provided in Table~\ref{bias}. The Bland-Altman plots and scatter plots are provided in Figure~\ref{B-A&Scatter}.

\begin{table}[H] 
\caption{Analysis on bias between PolyCam measurements and Fit3D measurements.\label{bias}}
\isPreprints{\centering}{} 
\begin{tabular}{llllllll}
\toprule
          & bias mean & bias std &  RMSE  & MAE  & ICC  & Pearson & p-value \\
\midrule
calf    & -15.02    & 8.13     & 17.08 & 15.72  & 0.926 & 0.984   & \textless{}0.001    \\
calf-1  & -15.67    & 7.63     & 17.43 & 16.01  & 0.931 & 0.989   & \textless{}0.001    \\
calf-2  & -14.34    & 8.56     & 16.70 & 15.42  & 0.918 & 0.978   & \textless{}0.001    \\
hip       & -12.92    & 13.96    & 19.02 & 14.31  & 0.984 & 0.992   & \textless{}0.001    \\
hip-1     & -13.53    & 13.34    & 19.00 & 14.40  & 0.982 & 0.991   & \textless{}0.001    \\
hip-2     & -12.27    & 14.55    & 19.04 & 14.21  & 0.986 & 0.992   & \textless{}0.001    \\
waist     & -7.82     & 10.16    & 12.81 & 10.29  & 0.994 & 0.996   & \textless{}0.001    \\
waist-1   & -8.98     & 8.88     & 12.63 & 10.44  & 0.992 & 0.996   & \textless{}0.001    \\
waist-2   & -6.65     & 11.17    & 13.00 & 10.14  & 0.992 & 0.994   & \textless{}0.001    \\
wrist   & -10.21    & 9.81     & 14.16 & 11.45 & 0.832 & 0.911   & \textless{}0.001    \\
wrist-1 & -10.98    & 10.63    & 15.28 & 12.14 & 0.814 & 0.906   & \textless{}0.001    \\
wrist-2 & -9.42     & 8.81     & 12.90 & 10.74 & 0.844 & 0.920    & \textless{}0.001   \\
\bottomrule
\end{tabular}
\flushleft
\noindent{\footnotesize{Note: The unit is mm; the p-value is calculated using the Wilcoxon signed-rank test; the "-1" or "-2" after location names means this group of measurements are from scans obtained in participants' 1st visit between 18-24 gestational weeks or the 2nd visit between 31-38 gestational weeks.}}
\end{table}

The Table~\ref{bias} shows that PolyCam consistently produced lower circumference measurements than Fit3D across all anatomical landmarks. When measuring larger circumferences such as hip and waist circumferences, PolyCam shows higher relative accuracy than on calf and wrist circumferences. For all landmarks, ICC values exceeded 0.80 at all landmarks and were greater than 0.90 for the calf, hip, and waist. And Pearson coefficients over 0.9 are observed, showing high agreements between two scanning methods. Except measurements on wrist, all other measurements have ICC over 0.9, which is a widely accepted threshold in clinical practice \cite{portney2009foundations}. Wilcoxon signed-rank test on all measurement pairs yielded $p < 0.001$. The RMSE and MAE scores lower than 20 mm, with RMSE ranging from 12.63 mm to 19.04 mm and MAE ranging from 10.14 mm to 16.01 mm. The Bland-Altman plots and scatter plots in Figure~\ref{B-A&Scatter} visualize the alignments and deviations between the two type of measurements. 

In Table~\ref{bias}, for the measurements extracted from scans of different pregnancy stages, comparable trends were observed across visits for both scanning methods. Along with Table~\ref{stat}, we notice that the bias is proportional to the length of circumferences within landmarks. In late pregnancy, participants have larger measurements across all landmarks, when bias between PolyCam and Fit3D are smaller.

\begin{figure}[H]
\isPreprints{\centering}{} % Only used for preprints
\includegraphics[width=\textwidth]{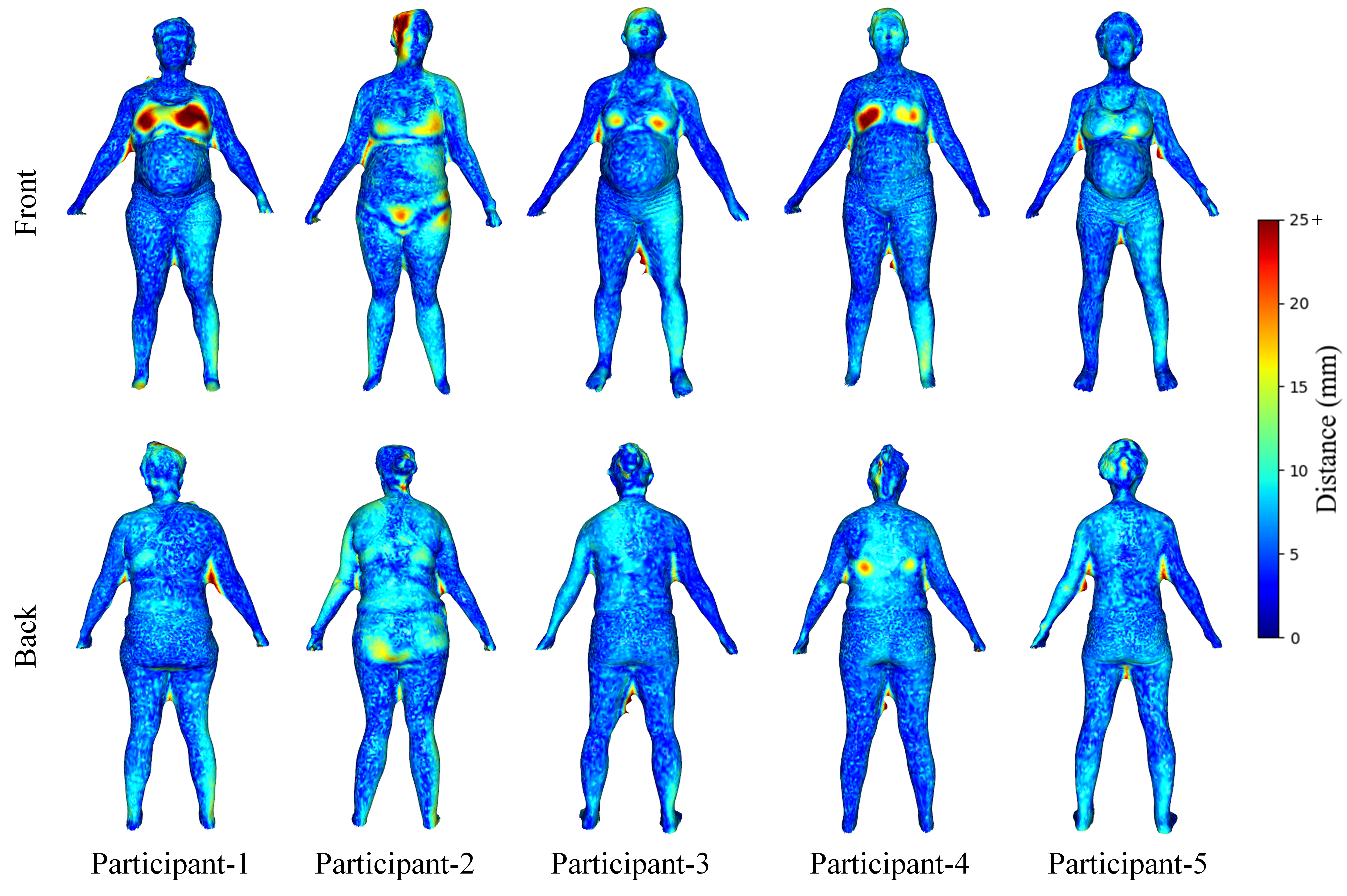}
\caption{Heatmaps of vertex-level deviations between registered PolyCam and Fit3D scans for five randomly selected participants from front view and back view. Colors represent the Euclidean distance between each PolyCam vertex and its nearest vertex on the corresponding registered Fit3D mesh, with blue indicating small deviations and red indicating large deviations. \label{heatmap}}
\end{figure}

To present a global view of scanning bias between the two methods, five representative participants were randomly selected and the heatmaps were generated to show the vertices-level bias from front and back views. In Figure~\ref{heatmap}, we observe that most of body regions across all examples exhibited vertex-level deviations below 5 mm. Larger deviations were primarily observed on breasts, underarms, and crotch regions, where bias larger than 25 mm can be found.

\subsection{Consistency Evaluation}
\begin{figure}[H]
\isPreprints{}{
\begin{adjustwidth}{-\extralength}{0cm}
\centering
}
\subfloat[\centering]{\includegraphics[width=5.0cm]{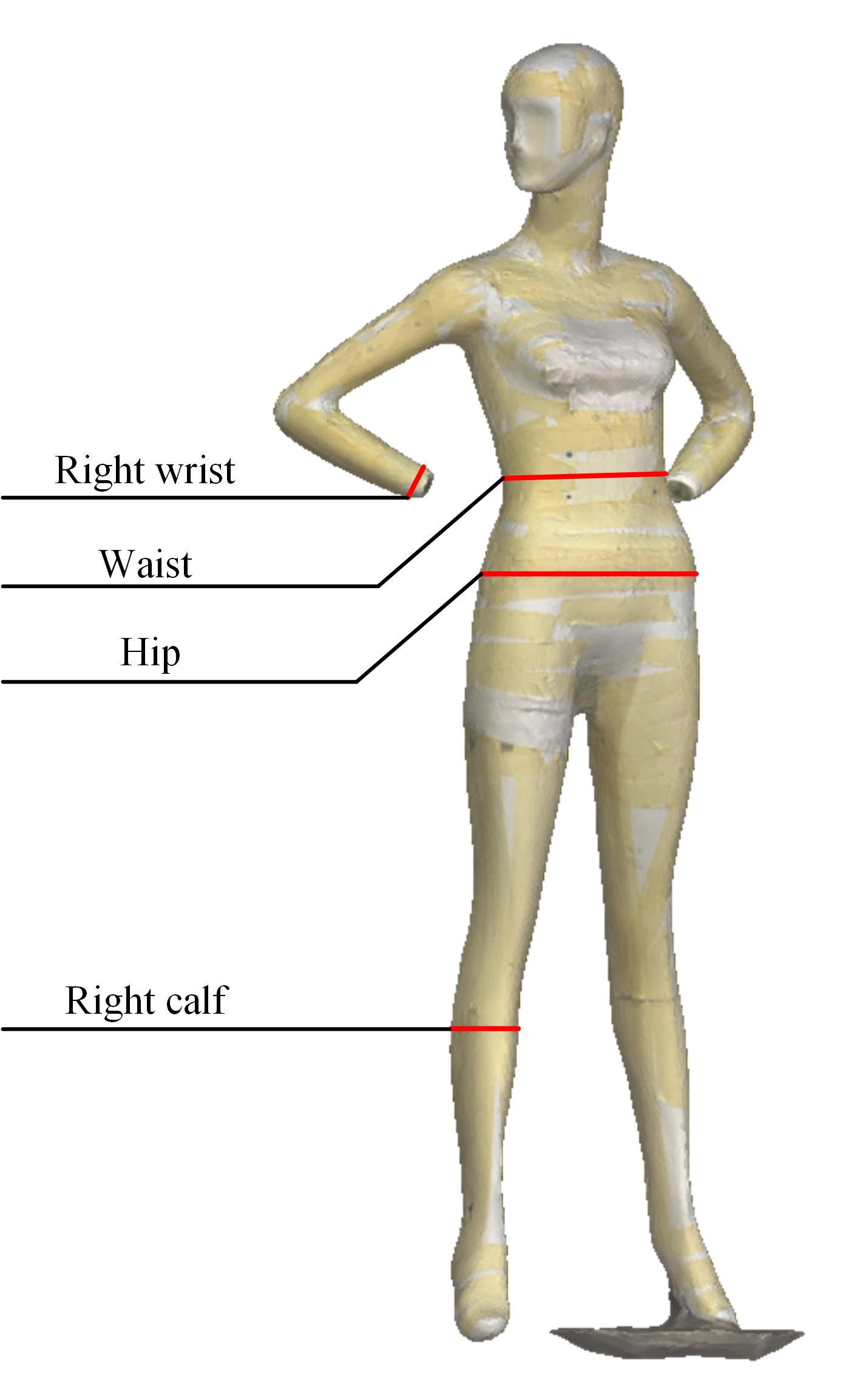}}
%\hfill
\subfloat[\centering]{\includegraphics[width=11.0cm]{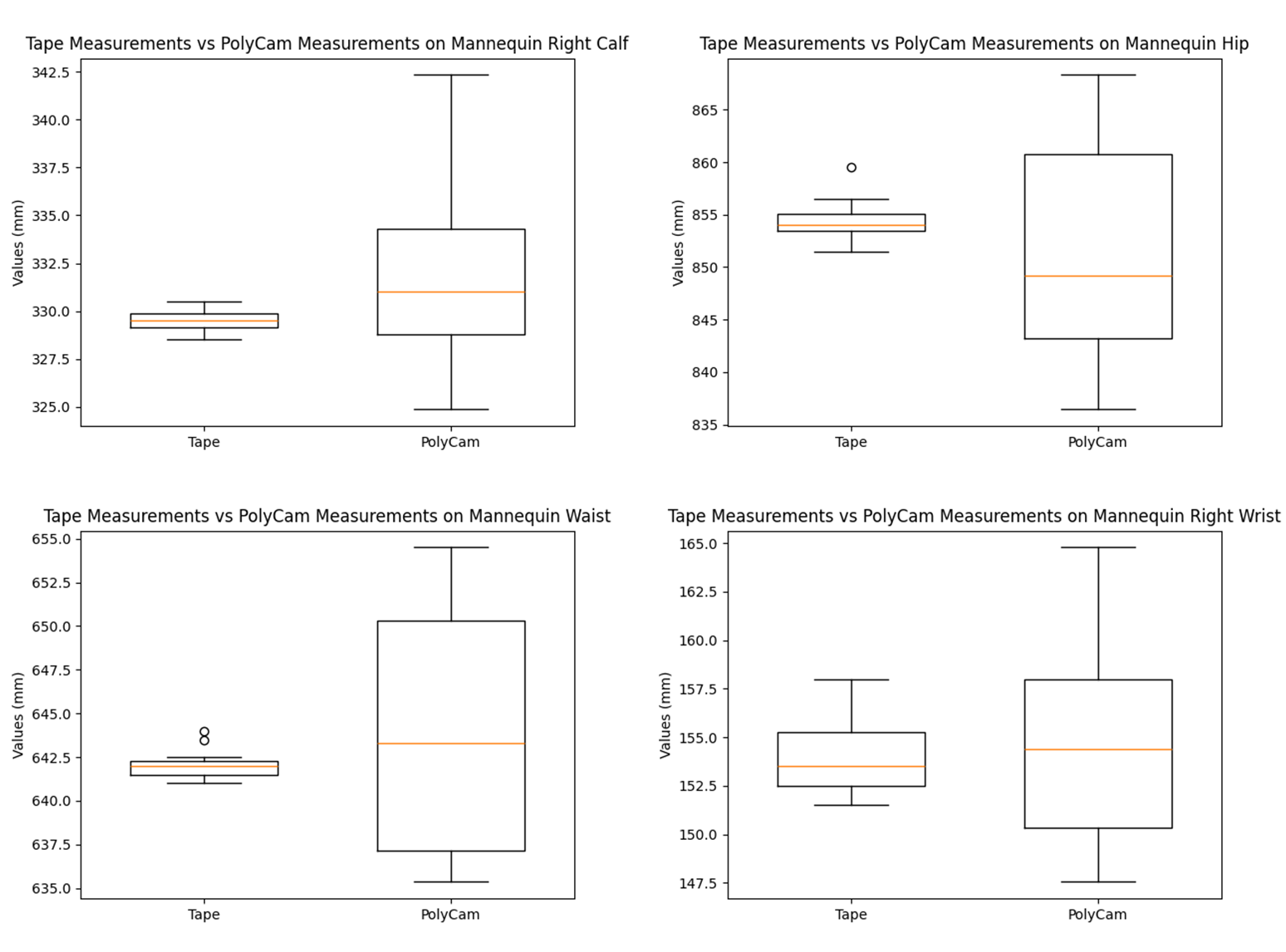}}\\
\isPreprints{}{
\end{adjustwidth}
} 
\caption{Landmarks marked on the rigid mannequin and distributions of repeated measurements obtained using PolyCam and tape measurements. (\textbf{a}) PolyCam scan of the rigid mannequin with landmarks highlighted for digital measurement. (\textbf{b}) Box plots of the 15 repeated measurements by tape and PolyCam each on the 4 landmarks.\label{mannequinplots}}
\end{figure}

The repeat PolyCam scans and tape measuring were conducted to get two groups of measurements on the 4 landmarks of the rigid mannequin. Since inevitable physiological motions such as breath are eliminated on the mannequin, the results can accurately reflect the capability of PolyCam and ensure the same distribution of samples for consistency evaluation. The landmarks on the mannequin are defined by the reference markers on the mannequin surface, as shown in Figure~\ref{mannequinplots}(\textbf{a}). Summary of the measurements from the two sources and analysis results are listed in Table~\ref{mannequinstat}. Box plots are presented for comparison in Figure~\ref{mannequinplots}(\textbf{b}).

\begin{table}[H]
\caption{Comparison between tape measurements and PolyCam-based measurements on a rigid mannequin.\label{mannequinstat}}
\isPreprints{\centering}{
	\begin{adjustwidth}{-\extralength}{0cm}
}
\isPreprints{\begin{tabularx}{\textwidth}{llllllllllll}}{
		\begin{tabularx}{\fulllength}{llllllllllll}
}
			\toprule
			& \multicolumn{5}{c}{PolyCam (mm) (n=15)}           & \multicolumn{5}{c}{Tape (mm) (n=15)}
            & \multicolumn{1}{c}{\multirow{2}{2.5cm}{Mann-Whitney U test p-value}} \\
            \cmidrule(r){2-6} \cmidrule(l){7-11}
             & mean    & std & min \%  & max \% & CV & mean    & std & min \%  & max \% & CV &  \\
			\midrule
calf  & 332.21 & 4.59  & -1.42\% & 3.88\% & 0.014    & 329.55 & 0.58  & -0.32\% & 0.29\% & 0.002    & 0.135                                             \\
hip      & 850.92 & 9.47  & -2.10\% & 1.63\% & 0.011    & 854.38 & 1.95  & -0.34\% & 0.60\% & 0.002    & 0.263                                             \\
waist    & 644.27 & 6.58  & -1.04\% & 1.94\% & 0.010    & 642.07 & 0.80  & -0.17\% & 0.30\% & 0.001    & 0.213                                             \\
wrist & 154.65 & 4.59  & -4.25\% & 6.92\% & 0.030    & 154.11 & 2.17  & -1.70\% & 2.52\% & 0.014    & 0.836\\
			\bottomrule
		\end{tabularx}
		\isPreprints{}{
	\end{adjustwidth}
}
\flushleft
\noindent{\footnotesize{Note: "min\%" is calculated by $(x_{min}-\bar{x}_{tape})/\bar{x}_{tape}$, where $x_{min}$ is the minimum measurement within the group and $\bar{x}_{tape}$ is the corresponding mean tape measurement; vice versa for "max\%".}}
\end{table}

From the results, we observe that on the rigid mannequin, where confounding factors associated with subject motion were eliminated, the bias of means are below 3.5 mm. Although both Table~\ref{mannequinstat} and Figure~\ref{mannequinplots}(\textbf{b}) show that PolyCam measurements have larger standard deviations than tape measurements, the relative extreme values and CVs show a low measurement variability of PolyCam measurements, with CV below 0.030. And by applying Mann-Whitney U test on the measurements from two sources, no significant differences were observed between PolyCam and tape measurements at any landmark.

%%%%%%%%%%%%%%%%%%%%%%%%%%%%%%%%%%%%%%%%%%
\section{Discussion}
This study evaluated the accuracy and consistency of a smartphone-based photogrammetric approach for whole-body 3D scanning by comparing PolyCam reconstructions with scans obtained using a commercial depth-sensor-based body scanner. Visual inspection of the reconstructed meshes indicated that PolyCam was capable of preserving the detailed whole-body surface geometry. Across the four automatically extracted circumference measurements, PolyCam showed strong agreement with Fit3D, with ICC values exceeding 0.80 and Pearson correlation coefficients exceeding 0.90 at all landmarks. Furthermore, comparable increases in circumference were also observed between the two pregnancy visits, indicating that both methods captured similar longitudinal body-shape changes. Repeated measurements of the rigid mannequin further showed good accuracy and consistency under controlled conditions.

Despite the strong agreement observed between the two methods, PolyCam measurements were consistently smaller than their corresponding Fit3D measurements. The mean differences (PolyCam - Fit3D) were -15.02 mm for the calf, -12.92 mm for the hip, -7.82 mm for the waist, and -10.21 mm for the wrist. However, PolyCam scans on the rigid mannequin did not show the consistent shrink on all measurements. This observation reveals the challenge of accurately scaling 3D body scans obtained by photogrammetric scanning on the non-rigid and rich-detailed human body. 

Additionally, although these differences represented a relatively small proportion of the measured circumferences, particularly for the hip and waist, previous studies of localized body regions have reported smaller measurement errors with PolyCam. Walters et al. reported RMSE values ranging from 1.99 mm to 2.31 mm for circumference measurements of reconstructed residual limbs, while Rudari et al. reported a standard deviation of 0.48 mm for finger cross-sectional circumferences reconstructed using PolyCam photo mode \cite{walters2024smartphone,rudari2024accuracy}. However, these studies focused on isolated body parts and acquired images specifically targeting the region of interest. In contrast, the present study evaluated whole-body scanning under realistic conditions, where inevitable postural sway and subtle body movements occur during image acquisition. These methodological differences likely contributed to the larger measurement deviations observed in the present study and should be considered when comparing the results with those reported for localized body-part reconstruction.

The mannequin experiment provides additional support for this interpretation. When subject motion was eliminated, the absolute difference between the mean PolyCam and mean reference tape measurements was below 3.5 mm at all four landmarks, approaching the magnitude of measurement errors reported in previous studies of localized body-part reconstruction \cite{walters2024smartphone}. Although PolyCam measurements exhibited greater variability than tape measurements, the Mann–Whitney U test did not detect statistically significant differences between the two measurement distributions at any landmark. These findings suggest that factors associated with whole-body acquisition, including subject motion and fixed view point, may contribute to the larger measurement deviations observed in human participants in addition to errors inherent to the photogrammetric reconstruction process.

\begin{figure}[H]
%\isPreprints{\centering}{} % Only used for preprints
\includegraphics[width=\textwidth]{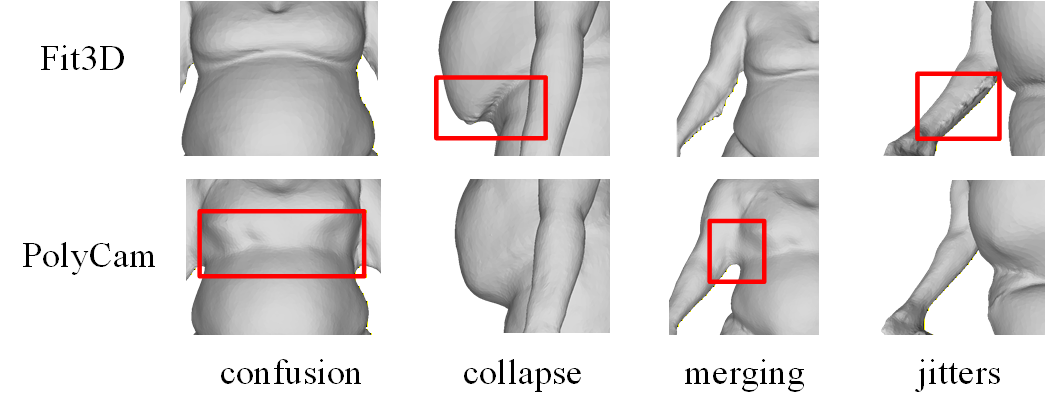}
\caption{Representative examples of reconstruction artifacts observed in PolyCam and Fit3D scans. Affected regions are highlighted by red rectangles. \label{examples}}
\end{figure}  

To better understand the observed measurement deviations, scan pairs exhibiting relatively large bias were further examined. Three primary factors were identified as potential sources of error. First, local reconstruction errors may arise from insufficient or ambiguous visual texture information. Heatmap analysis revealed that larger deviations frequently occurred in the breast region. During data collection, many participants wore dark-colored sports bras, which reduced visible shading and texture cues in RGB images. Because photogrammetric reconstruction relies on feature correspondence across multiple images, limited surface texture may reduce reconstruction accuracy and make it more difficult to distinguish local concave and convex geometry. Accordingly, the larger vertex-level deviations in these regions, as shown in Figure~\ref{examples}, may reflect local reconstruction artifacts in addition to differences between measurement systems. Second, the fixed image acquisition view point may have contributed the proportional distortion. In the present study, the smartphone remained fixed on a tripod at approximately waist height throughout scanning. Due to the perspective angle, body regions closer to the camera, such as the waist and hips, may have been reconstructed more accurately than regions farther from the optical center. This explanation is also compatible with the more proportional bias observed across landmarks in both the mannequin experiment and the longitudinal human scans, as well as with the spatial distribution of vertex-level deviations shown in the heatmaps, although the effect of camera position needs to be evaluated independently in future studies. (3) The scanning artifacts were also observed in some Fit3D scans. This issue brought about the outliers described in subsection 3.1. Surface collapse in pendulous regions, merging of adjacent body surfaces, and localized surface jitter were observed in several scans, as illustrated in Figure~\ref{examples}. Despite the use of three vertically aligned depth cameras, occlusion in regions such as the lower abdomen and hips may result in incomplete depth information and localized surface collapse. 

In addition to the systematic factors discussed above, several limitations of the current study design are listed below. First, while the present work compared two representative technologies, namely a photogrammetry-based smartphone application and a commercial depth-sensor-based body scanner, direct tape measurements were not included in the human participant study. Therefore, the comparison primarily reflects agreement between the two scanning approaches rather than absolute accuracy in human subjects. Second, the consistently negative bias observed across landmarks suggests a potential scaling issue in the photogrammetric reconstruction process. Although scaling was corrected using the width between predefined reference points and fine-tuned with ICP algorithm, residual scaling errors may still have contributed to the systematic underestimation of circumference measurements.

Future studies may further improve whole-body photogrammetric scanning by optimizing the study protocol. Potential strategies include the use of bright, form-fitting clothing with distinctive patterns to enhance feature matching, incorporating an external physical scale reference to improve scaling accuracy, and the acquisition of images obtained from multiple heights and viewing angles while maintaining a stationary subject position. Moreover, direct comparisons between tape measurements and photogrammetric reconstructions in human participants would provide a more comprehensive assessment of measurement accuracy.

With continued validation and optimization, photogrammetry-based whole-body scanning has the potential to provide a practical and accessible solution for longitudinal body-shape monitoring outside clinical environments. As a low-cost and non-invasive technology, it could facilitate telehealth, at-home self-monitoring, and community-based health assessment. Beyond conventional anthropometric measurements, photogrammetric whole-body scans preserve rich 3D surface geometry that may serve as a novel source of information for health assessment. These high-dimensional body shape data could complement traditional clinical variables and support the development of models for longitudinal phenotyping, disease screening, and prediction of conditions in which body morphology is clinically relevant. Beyond healthcare, photogrammetric body scanning also has potential applications in personalized garment design, virtual fitting, ergonomic assessment, and medico-legal documentation.

\section{Conclusion}
This study evaluated a smartphone-based photogrammetric approach for whole-body 3D scanning by comparing PolyCam-derived circumference measurements with those simultaneously obtained using the depth-sensor-based Fit3D ProScanner, and with tape measurements on a rigid mannequin. PolyCam showed strong agreement with Fit3D across the four evaluated landmarks and captured longitudinal changes in body circumference across pregnancy visits. Repeated mannequin measurements further demonstrated good absolute accuracy, repeatability and consistency under controlled conditions. These findings support the potential of smartphone-based photogrammetry as an accessible approach for whole-body anthropometric assessment and longitudinal body-shape monitoring. Further validation using independent scale references and direct anthropometric measurements in human participants is warranted before broader clinical or remote-monitoring applications are considered.

\vspace{6pt} 

\authorcontributions{For research articles with several authors, a short paragraph specifying their individual contributions must be provided. The following statements should be used ``Conceptualization, R.C. and J.H.; methodology, R.C. and B.F.; software, R.C.; validation, R.C. and B.F.; formal analysis, R.C. and Q.P.; investigation, R.C. and B,F.; resources, J.H.; data curation, R.C., and J.C.; writing---original draft preparation, R.C.; writing---review and editing, R.C., B.F., J.C., C.Q., Y.L., Q.P. and J.H.; visualization, R.C.; supervision, J.H.; project administration, J.H.; funding acquisition, J.H. }

\funding{Research reported in this publication was supported by the National Institute of Diabetes And Digestive and Kidney Diseases under Award Number R01DK129809 and by the National Institute on Aging under Award Number R56AG089080 of the National Institutes of Health. The content is solely the responsibility of the authors and does not necessarily represent the official views of the National Institutes of Health.}

\institutionalreview{The study was conducted in accordance with the Declaration of Helsinki, and approved by the Institutional Review Board of George Washington University (NCR224227, accepted on 08/31/2022).}

\informedconsent{Informed consent was obtained from all subjects involved in the study. All data were de-identified prior to analysis.}

\dataavailability{Data may be made available based on the nature of the request.}

\acknowledgments{We would like to express our sincere gratitude to all the volunteers who contributed their data to this study.}

\conflictsofinterest{The authors declare no conflicts of interest.} 

\reftitle{References}

\PublishersNote{}
\isPreprints{}{}

\end{document}